\documentclass{article}

\usepackage{graphicx} 
\usepackage{bold-extra}
\usepackage{subfigure} 
\usepackage{amsmath} 
\usepackage{amsfonts} 
\usepackage[bottom]{footmisc}
\usepackage{lmodern}
\newtheorem{theorem}{Theorem}[section]

\newtheorem{proposition}[theorem]{Proposition}

\newenvironment{sketch_proof}[1][Sketch of Proof]{\begin{trivlist}
\item[\hskip \labelsep {\bfseries #1}]}{\end{trivlist}}

\newcommand{\qed}{\nobreak \ifvmode \relax \else
      \ifdim\lastskip<1.5em \hskip-\lastskip
      \hskip1.5em plus0em minus0.5em \fi \nobreak
      \vrule height0.75em width0.5em depth0.25em\fi}

\usepackage{natbib}

\usepackage{algorithm}
\usepackage{algorithmic}

\usepackage{hyperref}

\usepackage[accepted]{icml2013} 

\icmltitlerunning{Maxout Networks}

\begin{document} 

\twocolumn[
\icmltitle{Maxout Networks}

\icmlauthor{Ian J. Goodfellow}{goodfeli@iro.umontreal.ca}
\icmlauthor{David Warde-Farley}{wardefar@iro.umontreal.ca}
\icmlauthor{Mehdi Mirza}{mirzamom@iro.umontreal.ca}
\icmlauthor{Aaron Courville}{aaron.courville@umontreal.ca}
\icmlauthor{Yoshua Bengio}{yoshua.bengio@umontreal.ca}
\icmladdress{D\'{e}partement d'Informatique et de Recherche Op\'{e}rationelle,
Universit\'{e} de Montr\'{e}al\\
2920, chemin de la Tour,
Montr\'{e}al, Qu\'{e}bec, Canada, H3T 1J8}


\icmlkeywords{multilayer perceptron, convolutional network, classification, computer vision}

\vskip 0.3in
]

\begin{abstract}
We consider the problem of designing models to leverage a recently introduced
approximate model averaging technique called {\em dropout}. We define a simple
new model called {\em maxout}
(so named because its {\em out}put is the {\em max}
of a set of inputs, and because it is a natural companion to dropout)
designed to both facilitate optimization by
dropout and improve the accuracy of dropout's fast approximate model averaging
technique. We empirically verify that the model successfully accomplishes both
of these tasks.
We use maxout and dropout to demonstrate state of the art classification performance on four
benchmark datasets: MNIST, CIFAR-10, CIFAR-100, and SVHN.
\end{abstract} 

\section{Introduction}

Dropout \citep{Hinton-et-al-arxiv2012} provides an inexpensive and simple means of
both training a large ensemble of models that share parameters
and
approximately averaging together these models' predictions. Dropout applied to multilayer perceptrons
and deep convolutional networks has improved 
the state of the art
on tasks ranging from  audio classification to very large scale object recognition
\citep{Hinton-et-al-arxiv2012,Krizhevsky-2012}. While dropout is known to work well in practice,
it has not previously been demonstrated
to actually perform model averaging for deep architectures
\footnote{Between submission and publication of this paper, we have learned that \citet{Srivastava-master-small}
performed experiments on this subject similar to ours.}
. Dropout is generally viewed as
an indiscriminately applicable tool that reliably yields a modest improvement in performance when applied
to almost any model.

We argue that rather than using dropout as a slight performance enhancement applied to arbitrary models,
the best performance may be obtained by directly designing a model that enhances dropout's abilities
as a model averaging technique. Training using dropout differs
significantly from previous approaches such as ordinary stochastic gradient descent. Dropout is most effective 
when taking relatively large steps in parameter space. In this regime, each update can be seen as making a significant update to a different model on a different subset
of the training set. The ideal operating regime for dropout is when the overall training procedure resembles
training an ensemble with bagging under parameter sharing constraints. This differs radically from the ideal stochastic gradient operating regime
in which a single model makes steady progress via small steps.
Another consideration is that dropout model averaging is only an approximation when applied to deep models.
Explicitly designing models to minimize this approximation error may thus enhance dropout's performance as well.

We propose a simple model that we call {\em maxout} that has beneficial characteristics both for optimization
and model averaging with dropout. We use this model in conjunction with dropout
to set the state of the art on four benchmark datasets
\footnote{Code and hyperparameters available at
\url{http://www-etud.iro.umontreal.ca/~goodfeli/maxout.html}
}
.

\section{Review of dropout}

Dropout is a technique that can be applied to deterministic feedforward architectures that predict an output
$y$ given input vector $v$. These architectures contain a series of hidden layers $\mathbf{h}=\{h^{(1)}, \dots, h^{(L)}\}$.
Dropout trains an ensemble of models consisting of the set of all models that contain a subset of the 
variables in both $v$ and
$\mathbf{h}$. The same set of parameters $\theta$ is used to parameterize a family of distributions
$p(y \mid v ; \theta, \mu)$ where $\mu \in \mathcal{M}$ is a binary mask determining which variables to include in the model.
On each presentation of a training example, we train a different sub-model by following the gradient of
$\log p(y \mid v; \theta, \mu)$ for a different
randomly sampled $\mu$. For many parameterizations of $p$ (such as most multilayer perceptrons) the instantiation of different
sub-models $p(y \mid v; \theta, \mu)$ can be obtained by elementwise multiplication of $v$ and $\mathbf{h}$  with the mask $\mu$.
Dropout training is similar to bagging \citep{ML:Breiman:bagging}, where many different models are trained on different subsets of the
data. Dropout training differs from bagging in that each model is trained for only one step and all of the models share
parameters. For this training procedure to behave as if it is
training an ensemble rather than a single model, each update must have a large effect, so that it makes the
sub-model induced by that $\mu$ fit the current input $v$ well.

The functional form becomes important when it comes time for the ensemble to make a prediction by averaging together all the sub-models' predictions.
Most prior work on bagging averages with the arithmetic mean, but it is not obvious
how to do so with the exponentially many models trained by dropout. Fortunately, some model families yield an inexpensive {\em geometric} mean.
When $p(y \mid v ; \theta) = \text{softmax}(v^T W + b)$, the predictive distribution defined by renormalizing the
geometric mean of $p(y \mid v ; \theta, \mu)$ over $\mathcal{M}$ is simply given by $\text{softmax}(v^T W /2 + b)$.
In other words, the average prediction of exponentially many sub-models can be computed simply by running the full model
with the weights divided by 2. This result holds exactly in the case of a single layer softmax model.
Previous work on dropout applies the same scheme in deeper architectures, such as multilayer perceptrons, where the $W/2$ method is only an approximation to the geometric mean.
The approximation has not been characterized mathematically, but performs well in practice.

\section{Description of maxout}

\begin{figure}[ht]
\begin{center}
\centerline{\includegraphics[width=\columnwidth]{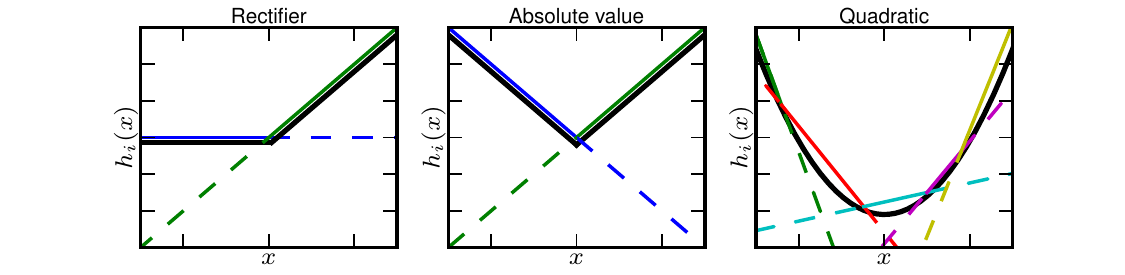}}
\caption{\small{Graphical depiction of how the maxout activation function can implement the 
rectified linear, absolute value rectifier, and approximate the quadratic activation function.
This diagram is 2D and
only shows how maxout behaves with a 1D input, but in multiple dimensions a maxout unit can
approximate arbitrary convex functions.}}
\label{piecewise}
\end{center}
\vskip -0.2in
\end{figure}

The maxout model is simply a feed-forward achitecture, such as a multilayer perceptron or deep convolutional
neural network, that uses a new type of activation function: the maxout unit. Given an input
$x \in \mathbb{R}^d$ ($x$ may be $v$, or may be a hidden layer's state), a maxout hidden layer implements the function
\[ h_i(x) = \max_{j \in [1,k]} z_{ij} \]
where  $z_{ij} = x^T W_{\cdots ij} + b_{ij}$, and $W \in \mathbb{R}^{d \times m \times k}$ and $b \in \mathbb{R}^{m \times k}$ are learned parameters. In a
convolutional network, a maxout feature map can be constructed by taking the maximum across
$k$ affine feature maps (i.e., pool across channels, in addition spatial locations).
When training with dropout, we perform the elementwise multiplication with the dropout mask immediately
prior to the multiplication by the weights in all cases--we do not drop inputs to the max operator.
A single maxout unit can be interpreted as making a piecewise
linear approximation to an arbitrary convex function. Maxout networks learn not just the relationship between hidden units, but also the
activation function of each hidden unit. See Fig. \ref{piecewise} for a graphical depiction
of how this works. 

\begin{figure}[ht]
\begin{center}
\centerline{\includegraphics[width=\columnwidth]{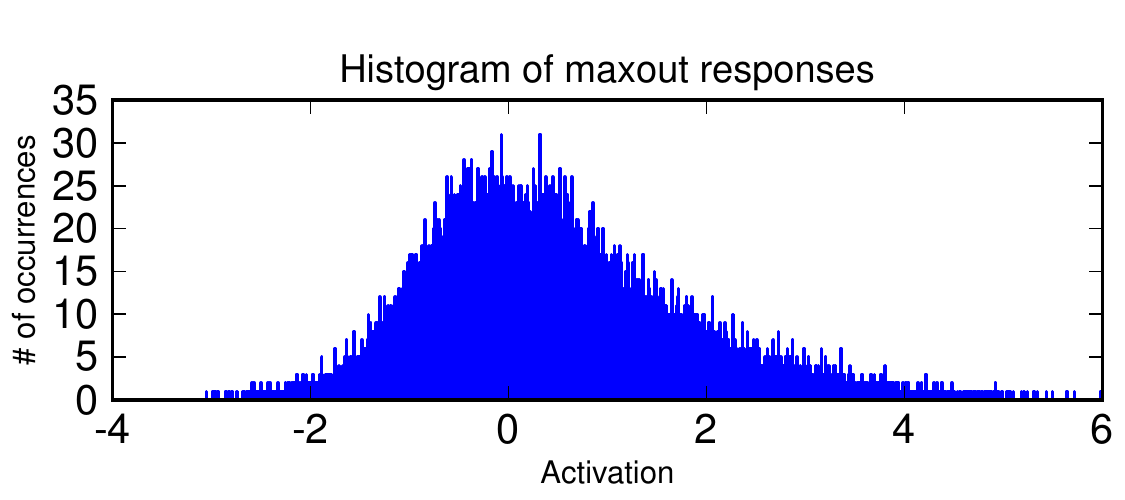}}
\caption{\small{The activations of maxout units are not sparse.}}
\label{not_sparse}
\end{center}
\end{figure}

Maxout abandons many of the mainstays of traditional activation function design. The representation
it produces is not sparse at all (see Fig. \ref{not_sparse}), though the gradient is highly sparse and dropout will artificially sparsify
the effective representation during training. While maxout may learn to saturate on one side or the other this
is a measure zero event (so it is almost never bounded from above).
While a significant proportion of parameter space corresponds to the function
being bounded from below, maxout is not constrained to learn to be bounded at all.
Maxout is locally linear almost everywhere, while
many popular activation functions have signficant curvature. Given all of these departures from
standard practice, it may seem surprising that maxout activation functions work at all, but we find
that they are very robust and easy to train with dropout, and achieve excellent performance.

\section{Maxout is a universal approximator}

\begin{figure}[ht]
\begin{center}
\centerline{\includegraphics[width=.75  \columnwidth]{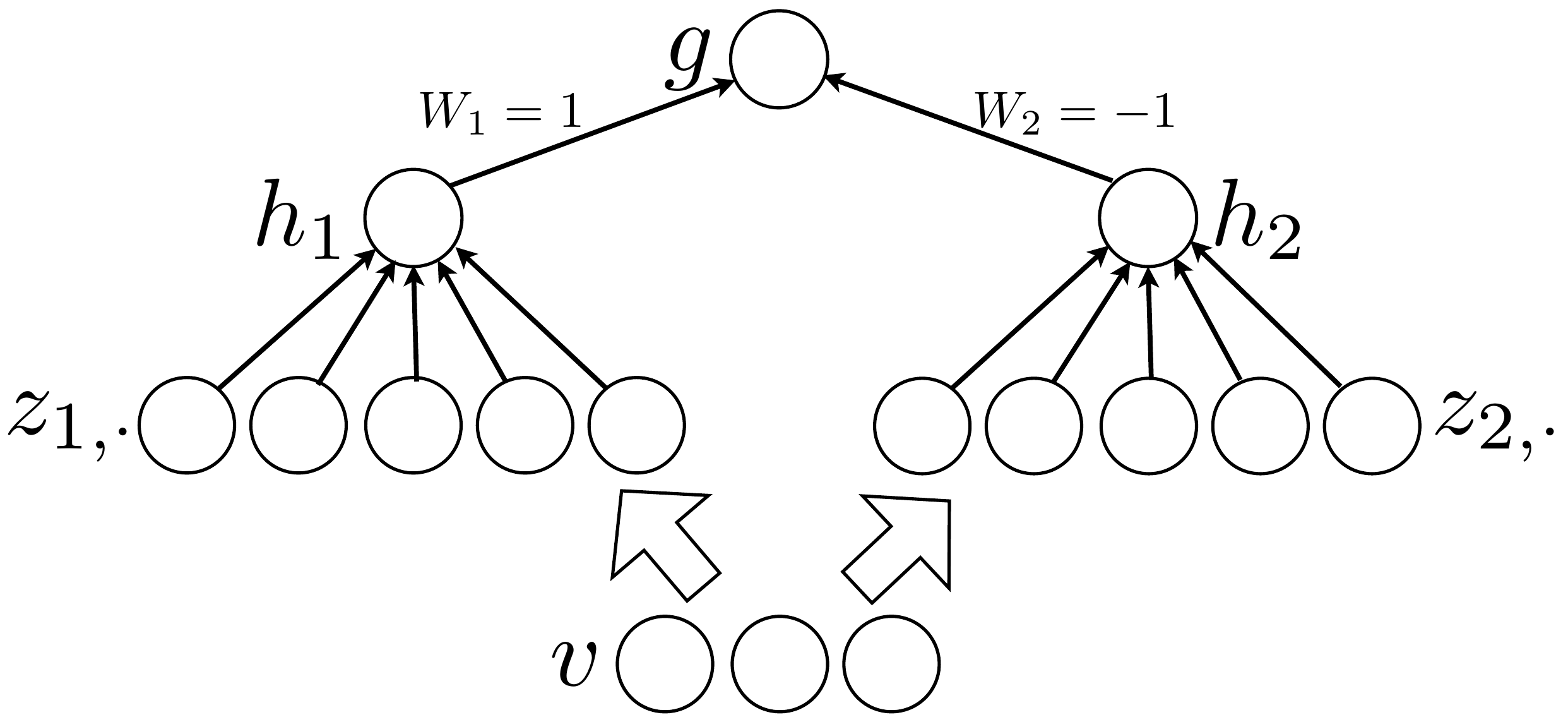}}
\caption{{\small An MLP containing two maxout units can arbitrarily approximate any continuous function.
The weights in the final layer can set $g$ to be the difference of $h_1$ and $h_2$. If $z_1$ and $z_2$ are allowed to have arbitrarily high cardinality, $h_1$ and $h_2$ can approximate
any convex function. $g$ can thus approximate any continuous function due to being a difference of
approximations of arbitrary convex functions.}}
\label{Uapprox}
\end{center}
\vskip -0.3in
\end{figure}

A standard MLP with enough hidden units is a universal approximator. Similarly, maxout networks are universal
approximators. Provided that each individual maxout unit may have arbitrarily many affine components,
we show that a maxout model with just two hidden units can approximate, arbitrarily well,
any continuous function of $v \in \mathbb{R}^n$.
A diagram illustrating the basic idea of the proof is presented in Fig. \ref{Uapprox}.


Consider the continuous piecewise linear (PWL) function $g(v)$ consisting of $k$ locally affine regions on
$\mathbb{R}^n$.

\begin{proposition} (From Theorem 2.1 in \cite{WangS2004})
For any positive integers $m$ and $n$, there exist two groups of $n+1$-dimensional real-valued parameter vectors
$[W_{1j},b_{1j}], j\in [1,k]$ and $[W_{2j},b_{2j}], j \in [1,k]$ such that:
\begin{equation}
g(v) = h_1(v) - h_2(v)
\end{equation}
That is, any continuous PWL function can be expressed as a difference of
two convex PWL functions. The proof is given in \cite{WangS2004}.
\label{prop1}
\end{proposition}

\begin{proposition} From the \emph{Stone-Weierstrass approximation theorem},
let
$C$ be a compact domain $C \subset \mathbb{R}^n$,
$f : C \rightarrow \mathbb{R}$ be a continuous function,
and $\epsilon > 0$ be any positive real number.
Then there exists a continuous PWL function $g$, (depending upon $\epsilon$),
such that for all $v \in C$, $\left| f (v) - g(v) \right| < \epsilon$.
\label{prop2}
\end{proposition}

\begin{theorem} \emph{Universal approximator theorem}. Any continuous function $f$ can be 
approximated arbitrarily well on a compact domain $C\subset \mathbb{R}^n$ by a maxout network
with two maxout hidden units.
\end{theorem}

\begin{sketch_proof}
  By Proposition \ref{prop2}, any continuous function can be
  approximated arbitrarily well (up to $\epsilon$), by a piecewise linear
  function. We now note that the representation of piecewise linear
  functions given in Proposition \ref{prop1} exactly matches a maxout network with two
  hidden units $h_1(v)$ and $h_2(v)$, with sufficiently large
  $k$ to achieve the desired degree
  of approximation $\epsilon$. Combining these, we conclude
  that a two hidden unit maxout network can approximate any
  continuous function $f(v)$ arbitrarily well on the compact domain $C$. In general as
  $\epsilon \rightarrow 0$, we have $k \rightarrow \infty$.
\end{sketch_proof}

\begin{figure}[ht]
\begin{center}
\centerline{\includegraphics[width= 0.5\columnwidth]{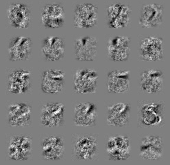}}
\caption{Example filters learned by a maxout MLP trained with dropout on MNIST. Each
row contains the filters whose responses are pooled to form a maxout unit.}
\label{mnist}
\end{center}
\vskip -0.4in
\end{figure} 

\section{Benchmark results}

\begin{table}[t]
\tiny
\caption{Test set misclassification rates for the best methods on the permutation invariant
MNIST dataset. Only methods that are regularized by modeling the input distribution outperform the maxout MLP.}
\label{MNIST-PI}
\vskip 0.15in
\begin{center}
\begin{small}
\begin{sc}
\begin{tabular}{p{5cm}c}
\hline
\abovespace\belowspace
Method & Test error \\
\hline
\abovespace
Rectifier MLP + dropout \citep{Srivastava-master-small} &1.05\%\\
DBM \citep{Salakhutdinov2009}    & 0.95\% \\
\textbf{Maxout MLP + dropout} & \textbf{ 0.94\%}\\
MP-DBM \citep{Goodfellow-arxiv2012} & 0.91\%\\
Deep Convex Network \citep{YuD11a} & 0.83\%\\
Manifold Tangent Classifier \citep{Dauphin-et-al-NIPS2011} & 0.81\%\\
DBM + dropout \citep{Hinton-et-al-arxiv2012} & 0.79\% \\
\hline
\end{tabular}
\end{sc}
\end{small}
\end{center}
\vskip -0.1in
\end{table}
We evaluated the maxout model on four benchmark datasets and set the state of the art on
all of them.

\subsection{MNIST}

The MNIST \citep{LeCun+98} dataset consists of 28 $\times$ 28 pixel greyscale images of handwritten
digits 0-9, with 60,000 training and 10,000 test examples. 
For the
{\em permutation invariant} version of the MNIST task, only methods unaware of the
2D structure of the data are permitted. For this task, we trained a model consisting of
two densely connected maxout layers followed by a softmax layer. We regularized the model
with dropout and by imposing a constraint on the norm of
each weight vector, as in \citep{Srebro05}. Apart from the maxout units, this is the same architecture used by
\citet{Hinton-et-al-arxiv2012}.
We selected the hyperparameters by minimizing the error on a validation set
consisting of the last 10,000 training examples.
To make use of the full training set, we recorded the value of the log likelihood on
the first 50,000 examples at the point of minimal validation error. We then continued
training on the full 60,000 example training set until the validation set log likelihood matched this
number. We obtained a test
set error of 0.94\%, which is the best result we are aware of that does not use unsupervised pretraining.
We summarize the best published results on permutation
invariant MNIST in Table \ref{MNIST-PI}.

We also considered the MNIST dataset without the permutation invariance restriction.
In this case, we used three convolutional maxout hidden layers (with spatial
max pooling on top of the maxout layers) followed by a densely connected softmax layer.
We were able to rapidly explore hyperparameter space thanks to the extremely fast GPU
convolution library developed by \citet{Krizhevsky-2012}.
We obtained a test set error rate of 0.45\%, which sets a new state of the art
in this category. (It is possible to get better results on MNIST by
augmenting the dataset with transformations of the standard set of images \citep{Ciresan-2010}
)
A summary of the best methods on the general MNIST dataset is provided
in Table \ref{MNIST}.

\begin{table}[t]
\tiny
\caption{Test set misclassification rates for the best methods on the general
MNIST dataset, excluding methods that augment the training data.}
\vskip 0.15in
\begin{center}
\begin{small}
\begin{sc}
\begin{tabular}{p{5cm}c}
\hline
\abovespace\belowspace
Method & Test error \\
\hline
\abovespace
2-layer CNN+2-layer NN \citep{Jarrett-ICCV2009} &0.53\%\\
Stochastic pooling  \cite{Zeiler+et+al-ICLR2013}  & 0.47\% \\
\textsc{\textbf{Conv. maxout + dropout}} & \textbf{ 0.45\%}\\
\hline
\end{tabular}
\end{sc}
\end{small}
\end{center}
\vskip -0.3in
\label{MNIST}
\end{table}

\subsection{CIFAR-10}

The CIFAR-10 dataset \citep{KrizhevskyHinton2009} consists of 32 $\times$ 32 color images drawn from
10 classes split into 50,000 train and 10,000 test images. We preprocess the data using global contrast normalization
and ZCA whitening.

We follow a similar procedure as with the MNIST dataset, with one change.
On MNIST, we find the best number of training epochs in terms of validation set error,
then record the training set log likelihood and continue training using the entire
training set until the validation set log likelihood has reached this value. On
CIFAR-10, continuing training in this fashion is infeasible because the final value of the
learning rate is very small and the validation set error is very high. Training until
the validation set likelihood matches the cross-validated value of the training likelihood
would thus take prohibitively long. Instead, we retrain the model from scratch, and stop
when the new likelihood matches the old one.

Our best model consists
of three convolutional maxout layers, a fully connected maxout layer, and
a fully connected softmax layer.
Using this approach we obtain a test set error of 11.68\%, which improves upon the
state of the art by over two percentage points.
(If we do not train on the validation set, we obtain a test set error of 13.2\%,
which also improves over the previous state of the art).
If we additionally augment the data with translations and horizontal reflections, we obtain the
absolute state of the art on this task at 9.35\% error. In this case, the likelihood during the
retrain never reaches the likelihood from the validation run, so we retrain for the same number
of epochs as the validation run.
A summary of the best CIFAR-10 methods is provided in Table \ref{CIFAR-10}.

\begin{table}[t]
\tiny
\caption{Test set misclassification rates for the best methods on the CIFAR-10 dataset.}
\label{CIFAR-10}
\vskip 0.15in
\begin{center}
\begin{small}
\begin{sc}
\begin{tabular}{p{5cm}c}
\hline
\abovespace\belowspace
Method & Test error \\
\hline
\abovespace
Stochastic pooling  \cite{Zeiler+et+al-ICLR2013}  & 15.13\% \\
CNN + Spearmint \cite{snoek-etal-2012b} & 14.98\%\\
\textbf{Conv. maxout + dropout} & {\bf 11.68 \%}\\
\hline
CNN + Spearmint + data augmentation \cite{snoek-etal-2012b} & 9.50 \% \\
\textbf{Conv. maxout + dropout + data augmentation} & {\bf 9.38 \%} \\
\end{tabular}
\end{sc}
\end{small}
\end{center}
\end{table}


\begin{figure}[ht]
\begin{center}
\centerline{\includegraphics[width=\columnwidth]{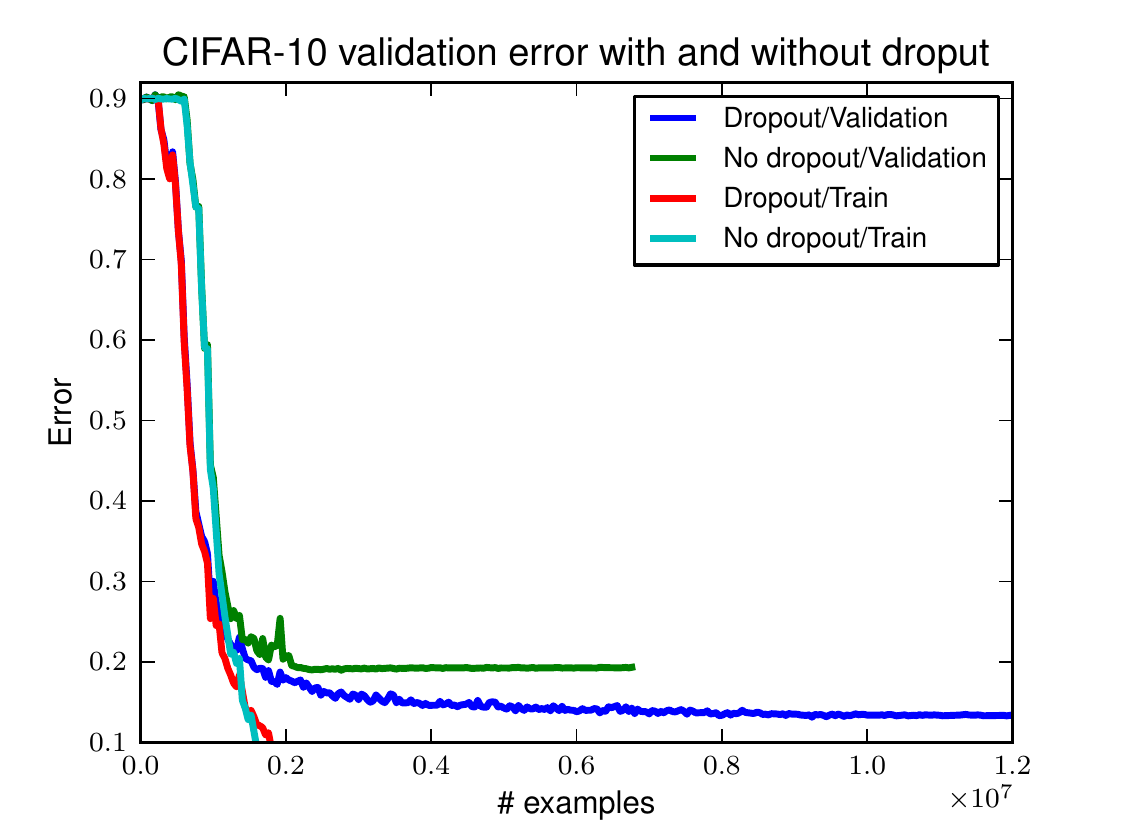}}
\caption{When training maxout, the improvement in validation set error that
results from using dropout is dramatic. Here we find a greater than 25\% reduction in our validation
set error on CIFAR-10.}
\label{perfect_storm}
\end{center}
\vskip -0.5in
\end{figure} 

\subsection{CIFAR-100}

The CIFAR-100 \citep{KrizhevskyHinton2009} dataset is the same size and format as the CIFAR-10 dataset,
but contains 100 classes, with only one tenth as many labeled examples per class.
Due to lack of time we did not extensively cross-validate hyperparameters
on CIFAR-100 but simply applied hyperparameters we found to work well on CIFAR-10.
We obtained a test set error of 38.57\%, which is state of the art.
If we do not retrain
using the entire training set, we obtain a test set error of 41.48\%,
which also surpasses the current state of the art.
A summary of the best methods on CIFAR-100 is provided in Table \ref{CIFAR-100}.

\begin{table}[t]
\tiny
\caption{Test set misclassification rates for the best methods on the CIFAR-100 dataset.}
\label{CIFAR-100}
\vskip 0.15in
\begin{center}
\begin{small}
\begin{sc}
\begin{tabular}{p{5cm}c}
\hline
\abovespace\belowspace
Method & Test error \\
\hline
\abovespace
Learned pooling \citep{MalinowskiFritzICLRwkshp2013} & 43.71\%\\
Stochastic pooling\cite{Zeiler+et+al-ICLR2013}    & 42.51\% \\
\textbf{Conv. maxout + dropout } & {\bf 38.57\%}\\
\hline
\end{tabular}
\end{sc}
\end{small}
\end{center}
\vskip -0.3in
\end{table}

\subsection{Street View House Numbers}

The SVHN \citep{Netzer-wkshp-2011} dataset consists of color images of house
numbers collected by Google Street View. The dataset comes in two formats.
We consider the second format, in which each image is of size 32 $\times$ 32
and the task is to classify the digit in the center of the image. Additional
digits may appear beside it but must be ignored. There are 73,257 digits in the training set, 26,032 digits in the test set
and 531,131 additional, somewhat less difficult examples, to use as an extra training set.
Following \citet{sermanet-icpr-12}, to build a validation set, we select 400 samples per class
from the training set and 200 samples per class from the extra set. 
The remaining digits of the train and extra sets are used for training.

For SVHN, we did not train on the validation set at all. We used
it only to find the best hyperparameters. We applied local contrast normalization preprocessing the same way
as \citet{Zeiler+et+al-ICLR2013}. Otherwise, we followed the same approach as on MNIST.
Our best model consists of three convolutional maxout hidden layers and a densely connected maxout layer followed by a densely connected softmax layer. We obtained a test set error rate of 2.47\%, which
sets the state of the art. A summary of comparable methods is provided in Table \ref{SVHN}.

\begin{table}[t]
\caption{Test set misclassification rates for the best methods on the SVHN dataset.}
\label{SVHN}
\vskip 0.15in
\begin{center}
\begin{small}
\begin{sc}
\begin{tabular}{p{5cm}c}
\hline
\abovespace\belowspace
Method & Test error \\
\hline
\abovespace
\cite{Sermanet-et-al-arxiv2012}& 4.90\%\\
Stochastic pooling  \cite{Zeiler+et+al-ICLR2013} & 2.80 \% \\
Rectifiers + dropout \cite{Srivastava-master-small}&  2.78 \% \\
Rectifiers + dropout + synthetic translation \cite{Srivastava-master-small} & 2.68 \% \\
\textbf{ Conv. maxout + dropout} & {\bf 2.47 \%}\\
\hline
\end{tabular}
\end{sc}
\end{small}
\end{center}
\vskip -0.3in
\end{table}

\section{Comparison to rectifiers}

One obvious question about our results is whether we obtained them by improved preprocessing or
larger models, rather than by the use of maxout. For MNIST we used no preprocessing, and for SVHN, we use
the same preprocessing as \citet{Zeiler+et+al-ICLR2013}. However on the CIFAR datasets we did use a new
form of preprocessing. We therefore compare maxout to rectifiers run with the same processing and a variety
of model sizes on this dataset.

\begin{figure}[ht]
\vskip -0.1in
\begin{center}
\centerline{\includegraphics[width=\columnwidth]{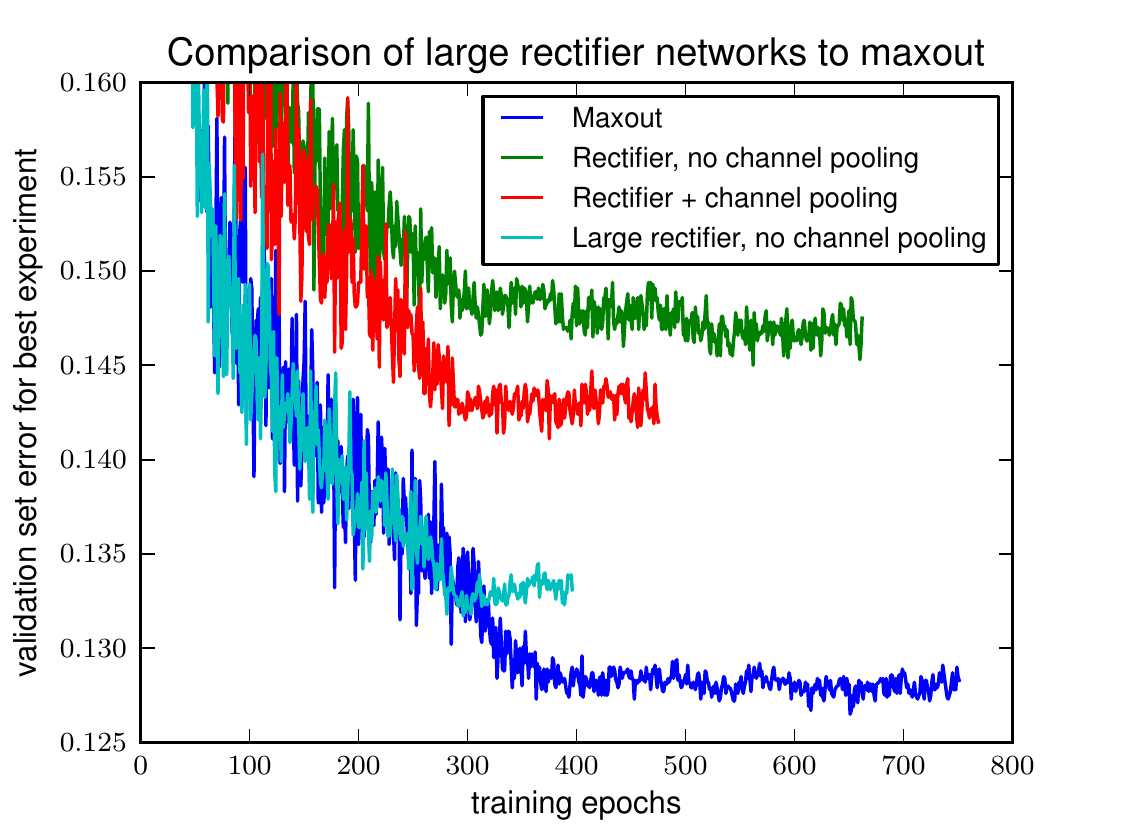}}
\caption{
We cross-validated the momentum and learning rate for four architectures of model:
1) Medium-sized maxout network.
2) Rectifier network with cross-channel pooling, and exactly the same number of
parameters and units as the maxout network.
3) Rectifier network without cross-channel pooling, and the same number of units as
		  the maxout network (thus fewer parameters).
4) Rectifier network without cross-channel pooling, but with $k$ times as many units as
the maxout network. Because making layer $i$ have $k$ times more outputs increases the
number of inputs to layer $i + 1$, this network has roughly $k$ times more parameters
than the maxout network, and requires significantly more memory and runtime.
We sampled 10 learning rate and momentum schedules and random seeds for dropout, then ran
each configuration for all 4 architectures. Each curve terminates after failing to improve the validation error in the last 100 epochs.
}
\label{size}
\end{center}
\vskip -0.2in
\vspace{-.2in}
\end{figure}

By running a large cross-validation experiment (see Fig. \ref{size}) we found that maxout offers a clear
improvement over rectifiers. We also found that our preprocessing and size of models improves rectifiers
and dropout beyond the previous state of the art result.
Cross-channel pooling is a method for reducing the size of state and number of parameters needed to have
a given number of filters in the model. Performance seems to correlate well with the number of filters for
maxout but with the number of output units for rectifiers--i.e, rectifier units do not benefit
much from cross-channel pooling. Rectifier units do best without cross-channel pooling but with the same
number of filters, meaning that the size of the state and the number of parameters must be about $k$ times
higher for rectifiers to obtain generalization performance approaching that of maxout.

\section{Model averaging}
\label{model_averaging}

\begin{figure}[ht]
\begin{center}
\centerline{\includegraphics[width=\columnwidth]{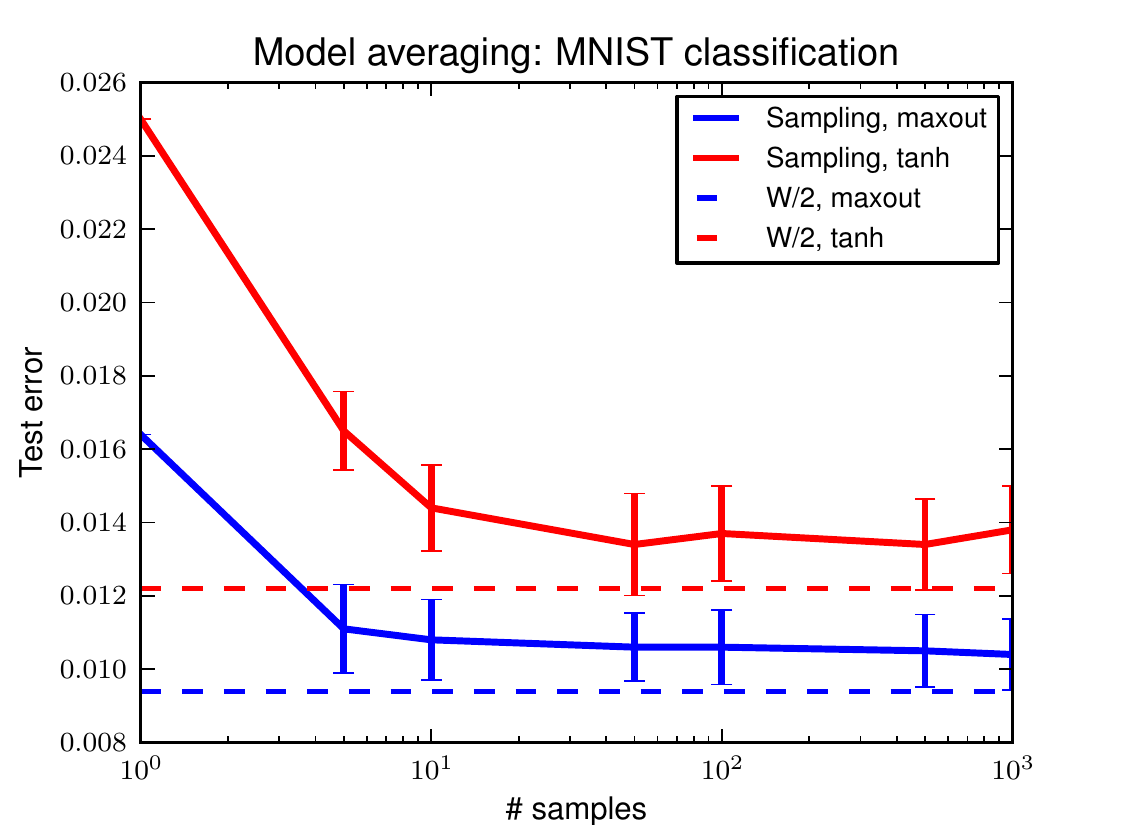}}
\caption{\small{The error rate of the prediction obtained by sampling several sub-models and taking the geometric mean of their predictions approaches the
error rate of the prediction made by dividing the weights by 2. However, the divided weights still
obtain the best test error, suggesting that dropout is a good approximation to averaging over a very
large number of models. Note that the correspondence is more clear in the case of maxout.}}
\label{sampling_error}
\end{center}
\vskip -0.4in
\end{figure} 

\begin{figure}[ht]
\begin{center}
\centerline{\includegraphics[width=\columnwidth]{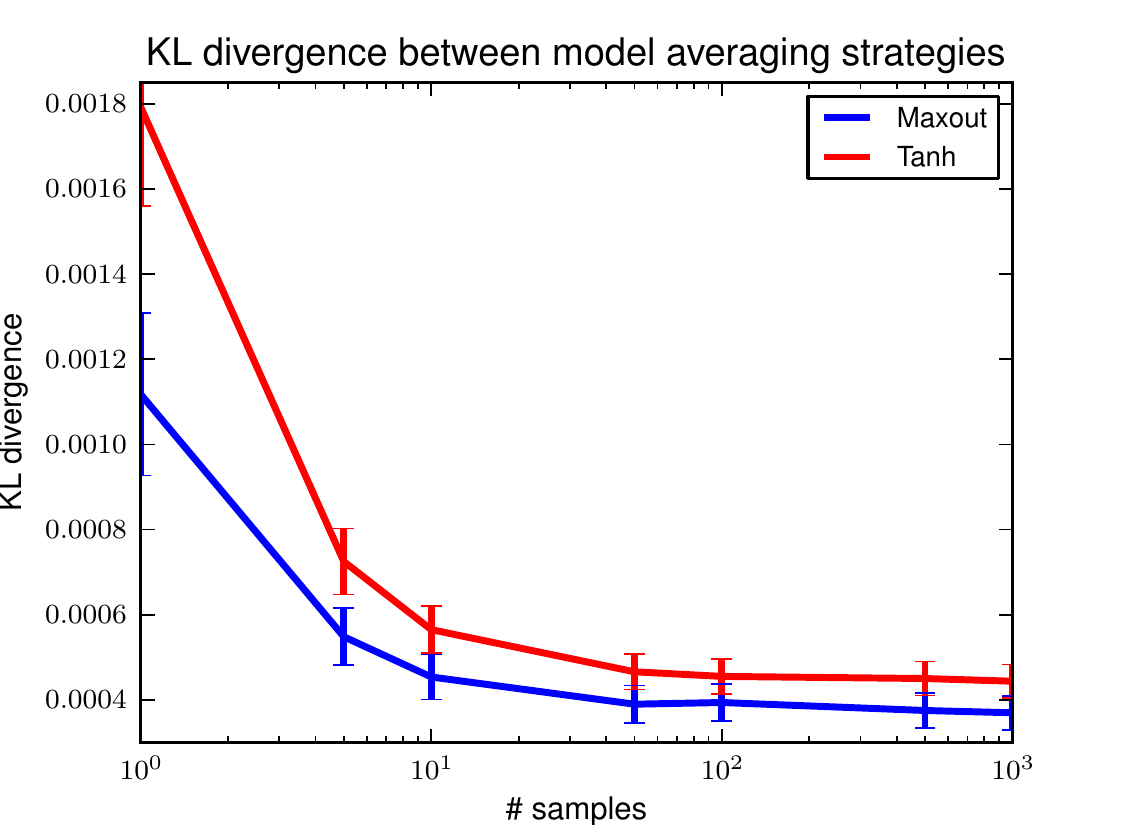}}
\caption{The KL divergence between the distribution predicted using the dropout technique of
dividing the weights by 2 and the distribution obtained by taking the geometric mean of the
predictions of several sampled models decreases as the number of samples increases. This suggests
that dropout does indeed do model averaging, even for deep networks. The approximation is more
accurate for maxout units than for tanh units.}
\vspace{-.51in}
\label{kl}
\end{center}
\end{figure} 

Having demonstrated that maxout networks are effective models, we now analyze
the reasons for their success. We first identify reasons that maxout is highly compatible
with dropout's approximate model averaging technique.

The intuitive justification for averaging sub-models by dividing the weights by 2
given by \citep{Hinton-et-al-arxiv2012} is that this does exact model averaging for a single layer model,
softmax regression. To this characterization, we add the observation that the model
averaging remains exact if the model is extended
to multiple linear layers. While this has the same representational power as a single layer,
the expression of the weights as a product of several matrices could have a different inductive
bias. More importantly, it indicates that dropout does exact model averaging in deeper architectures
provided that they are locally linear among the space of inputs to each layer that are visited
by applying different dropout masks.

We argue that dropout training encourages maxout units to have
large linear regions around inputs that appear in the training data. Because each sub-model
must make a good prediction of the output, each unit
should learn to have roughly the same activation regardless of which inputs are dropped. In a maxout network with arbitrarily selected parameters, varying the dropout mask will often
move the effective inputs far enough to escape the local region surrounding the clean inputs in which the hidden
units are linear, i.e., changing the dropout mask could frequently change which piece of the piecewise function
an input is mapped to. Maxout {\em trained with dropout} may have the identity of the maximal filter in
each unit change relatively rarely as the dropout mask changes. Networks of
linear operations and $\mathrm{max}(\cdot)$ may learn to exploit dropout's approximate model averaging technique
well.

Many popular activation functions have significant curvature nearly everywhere. These observations
suggest that the approximate model averaging of dropout will not be as accurate for networks incorporating such activation functions.
To test this, we compared the best maxout model trained on MNIST with dropout to a hyperbolic tangent
network trained on MNIST with dropout. We sampled several subsets of each model and compared the
geometric mean of these sampled models' predictions to the prediction made using the dropout technique
of dividing the weights by 2. We found evidence that dropout is indeed performing model averaging,
even in multilayer networks, and
that it is more accurate in the case of maxout. See Fig. \ref{sampling_error} and Fig. \ref{kl}
for details.

\section{Optimization}

The second key reason that maxout performs well is that it improves the bagging style training
phase of dropout.
Note that the arguments in section \ref{model_averaging} motivating the use of maxout also 
apply equally to rectified linear units \citep{Salinas1996, Hahnloser98, Glorot+al-AI-2011}.
The only difference between maxout and max pooling over a set of rectified linear
units is that maxout does not include a 0 in the max.
Superficially, this seems to be a small difference, but we find that including this constant
0 is very harmful to optimization in the context
of dropout. For instance, on MNIST our best validation set error with an MLP is 1.04\%. If we
include a 0 in the max, this rises to over 1.2\%. We argue that, when trained with dropout,
maxout is easier to optimize than rectified linear units with cross-channel pooling.

\subsection{Optimization experiments}

\begin{figure}[ht]
\vskip -0.1in
\begin{center}
\centerline{\includegraphics[width=\columnwidth]{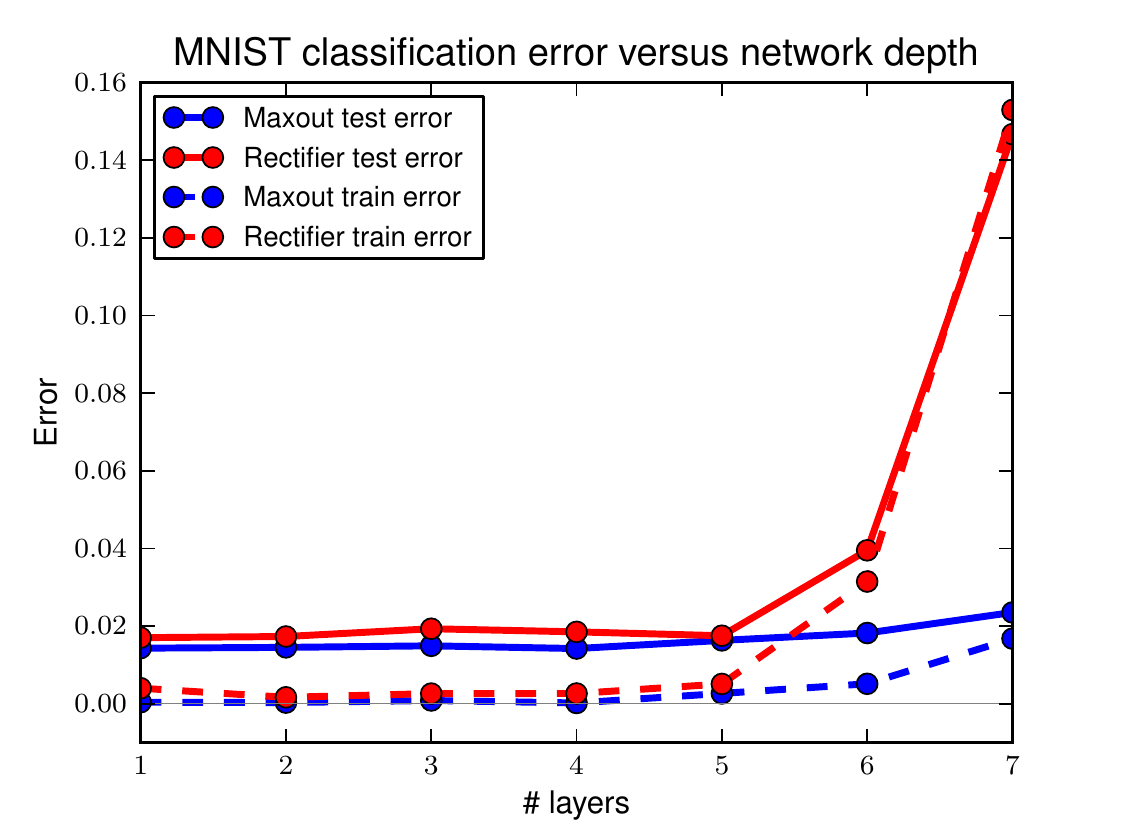}}
\caption{We trained a series of models with increasing depth on MNIST. Each layer contains only 80 units
($k$=5) to make fitting the training set difficult. Maxout optimization degrades gracefully
with depth but pooled rectifier units worsen noticeably at 6 layers and dramatically at 7.
}
\label{depth}
\end{center}
\vskip -0.3in
\end{figure}

To verify that maxout yields better optimization performance than max pooled rectified linear
units when training with dropout, we carried out two experiments. First, we stressed the optimization
capabilities of the training algorithm by training a small (two hidden convolutional layers with $k=2$
and sixteen kernels) model on the large (600,000 example) SVHN dataset. When training with rectifier
units the training error gets stuck at 7.3\%. If we train instead with maxout units, we obtain 5.1\% training error. 
As another optimization stress test, we tried training very deep and narrow models on MNIST, and found
that maxout copes better with increasing depth than pooled rectifiers. See Fig. \ref{depth} for details.

\subsection{Saturation}

\begin{figure}[ht]
\begin{center}
\centerline{\includegraphics[width=\columnwidth]{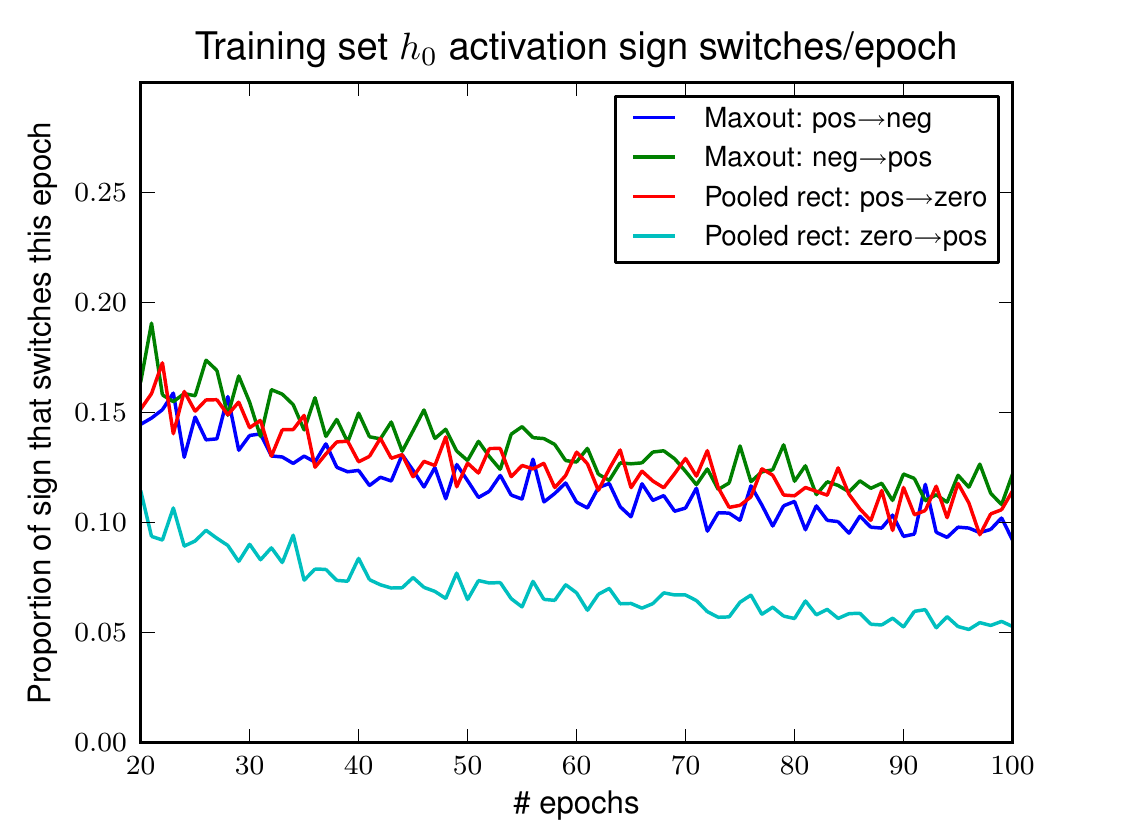}}
\caption{During dropout training, rectifier units transition from positive to 0 activation more
frequently than they make the opposite transition, resulting a preponderence of 0 activations.
Maxout units freely move between positive and negative signs at roughly
equal rates.}
\label{activation_switching}
\end{center}
\vskip -.4in
\end{figure} 

Optimization proceeds very differently when using dropout than when using ordinary stochastic gradient descent.
SGD usually works best with a small learning rate that results in a smoothly decreasing objective function,
while dropout works best with a large learning rate, resulting in a constantly fluctuating objective function.
Dropout rapidly explores many different directions and rejects the ones that worsen performance, while SGD
moves slowly and steadily in the most promising direction. We find empirically that these different operating regimes
result in different outcomes for rectifier units. When training with SGD, we find that the rectifier
units saturate at 0 less than 5\% of the time. When training with dropout, we initialize the units to sature rarely
but training gradually increases their saturation rate to 60\%. Because the 0 in the $\max(0,z)$ activation function is a constant, this blocks the gradient from flowing through the unit. In the absence of gradient
through the unit, it is difficult for training to change this unit to become active again. {\em Maxout does not
suffer from this problem because gradient always flows through every maxout unit}--even when a maxout unit is 0, this 0 is a function of the parameters and may be adjusted Units that take on negative
activations may be steered to become positive again later. Fig. \ref{activation_switching} illustrates how active rectifier units become
inactive at a greater rate than inactive units become active when training with dropout, but maxout units, which are
always active, transition between positive and negative activations at about equal rates in each direction.
We hypothesize that the high proportion of zeros and the difficulty of escaping them impairs the optimization
performance of rectifiers relative to maxout.

To test this hypothesis, we trained two MLPs on MNIST, both with two hidden layers and 1200 filters per layer pooled in groups of 5. When we include
a constant 0 in the max pooling, the resulting trained model fails to make use of 17.6\% of the filters in
the second layer and 39.2\% of the filters in the second layer. A small minority of the filters usually took
on the maximal value in the pool, and the rest of the time the maximal value was a constant 0. Maxout, on the
other hand, used all but 2 of the 2400 filters in the network. Each filter in each maxout unit in the network was maximal for some
training example. All filters had been utilised and tuned.

\subsection{Lower layer gradients and bagging}

To behave differently from SGD, dropout requires the gradient to change noticeably as the choice
of which units to drop changes. If the gradient is approximately constant with respect to the dropout mask,
then dropout simplifies to SGD training. We tested the hypothesis that rectifier networks suffer from diminished gradient
flow to the lower layers of the network by monitoring the variance with respect to dropout masks for fixed data during
training of two different MLPs on MNIST. The variance of the gradient on the output weights was 1.4 times
larger for maxout on an average training step, while the variance on the gradient of the first layer weights
was 3.4 times larger for maxout than for rectifiers. Combined with our previous result
showing that maxout allows training deeper networks, this 
greater variance suggests that maxout better propagates varying information
downward to the lower layers and helps dropout training to better resemble bagging for the lower-layer parameters. Rectifier networks,
with more of their gradient lost to saturation, presumably cause dropout training to resemble regular SGD toward the bottom of the network.

\vspace*{-2mm}
\section{Conclusion}
\vspace*{-1mm}

We have proposed a new activation function called maxout that is particularly well suited for
training with dropout, and for which we have proven a universal approximation theorem.
We have shown empirical evidence that dropout attains a good approximation
to model averaging in deep models. We have shown that maxout exploits this model averaging behavior because
the approximation is more accurate for maxout units than for tanh units.
We have demonstrated that optimization behaves very differently in the context of dropout than in the pure SGD case. By
designing the maxout gradient to avoid pitfalls such as failing to use many of a model's filters, we are able to
train deeper networks than is possible using rectifier
units. We have also shown that maxout propagates variations in the gradient due to different choices of dropout
masks to the lowest layers of a network, ensuring that every parameter in the model can enjoy the full
benefit of dropout and more faithfully emulate bagging training. The state of the art
performance of our
approach on five different benchmark tasks motivates the design of further models that are
explicitly intended to perform well when combined with
inexpensive approximations to model averaging.

\section*{Acknowledgements}

The authors would like to thank the developers of Theano \citep{bergstra+al:2010-scipy,Bastien-Theano-2012}, in particular Fr\'{e}d\'{e}ric Bastien and Pascal Lamblin for their assistance with infrastructure development and performance optimization. We would also like to thank Yann Dauphin for helpful discussions.

{\small
\bibliography{strings,strings-shorter,ml,aigaion-short}
\bibliographystyle{icml2013}
}

\end{document}